\documentclass[letterpaper]{article}
\usepackage{aaai}
\usepackage{times}
\usepackage{helvet}
\usepackage{courier}
\begin{document}
%

\title{Establishing Meta-Decision-Making for AI: An Ontology of
Relevance, Representation and Reasoning} 
\author{Cosmin Badea\textsuperscript{1}, Leilani Gilpin\textsuperscript{2}\\
 \textsuperscript{1}Department of Computing, Imperial College London, UK.\textsuperscript{2}University of California, Santa Cruz, United States.\\
 \emph{Contact information:} \textsuperscript{1}c.badea@imperial.ac.uk
Rule-based AI,
Anticipatory Thinking,
Cognitive Systems,
Third Wave Autonomous Systems, Risk Management\\
 }

\maketitle

\begin{abstract}
We propose an ontology of building decision-making systems, with the aim of establishing Meta-Decision-Making for Artificial Intelligence (AI), improving autonomy, and creating a framework to build metrics and benchmarks upon. To this end, we propose the three parts of Relevance, Representation, and Reasoning, and discuss their value in ensuring safety and mitigating risk in the context of third wave cognitive systems. Our nomenclature reflects the literature on decision-making, and our ontology allows researchers that adopt it to frame their work in relation to one or more of these parts.

\end{abstract}

\section{Meta-Decision-Making for AI}

Making good decisions is a very important part of constructing good Artificial Intelligence (AI). However, there is an important distinction between decision-making itself and reasoning about decision-making, similarly to the distinction between (normative) ethics and metaethics. We believe more focus in the areas of automated decision-making, anticipatory thinking and cognitive systems ought to be explicitly given to discussing and deciding upon the characteristics of good decision-making systems and how best to build them. We call this \textbf{meta-decision-making}. Meta-decision-making has been discussed before in domains such as medicine~\cite{medical} or action research~\cite{meta}. Herein we aim to help establish meta-decision-making in the context of AI as an important research avenue. 



The core of our contribution in this paper is a proposed ontology or classification of the different aspects of building a decision-making system. We identify key components in the current approaches and establish a nomenclature to 
help focus research efforts in the particular area that they are meant to aim at, improve our collective understanding of where a particular piece of work should fit in the process of building such a system and, most importantly, help develop autonomy capabilities of such agents. In particular, third wave autonomous systems, defined as context-adapting in \cite{DARPA}, should greatly benefit from this approach, as we try to illustrate below. Lastly, in order to measure success, we can use this ontology to build metrics and benchmarks, by focusing on a particular step of the system and auditing it, and this should also feed into the perception problem, as we illustrate with a self-driving car example.


\subsection{Literature - challenges}

Automated decision-making encapsulates many different modalities and parts, e.g., perception, sensors, planning, tactics, etc. When these systems fail, errors are generally dealt with in isolation (failure detection is done locally\cite{local-anomaly}) and thus issues can arise around not taking into account the connections and (logical) dependencies between the different parts of the system. 


Another issue, especially in autonomous systems, is that some decisions are safety-critical and time-sensitive. Reconciling inconsistencies is largely a manual process that can be time consuming. It also does not occur in real time, and is largely \emph{ex post facto} \cite{ai-failures}. 

One way to deal with or preempt failure in such a system is to use preferences and rule-based decision-making~\cite{preferences-come-from}. For example, in the field of moral reasoning, there is value-based decision-making with a rule-based implementation \cite{mars}. The focus of such works is generally on the preference ordering on the values (the Representation step we discuss below), or on the ordering on the rules (the Reasoning step below). We will use this implementation from~\cite{mars} as a running example below.

There are limitations to this approach, as well as the others found in the literature, especially in terms of \emph{adaptability} to new and unseen corner cases, \emph{anticipation} of new risks and failure cases, and in terms of issues of \emph{misinterpretation}~\cite{interpretation} of the symbolic representation of meaning. For example, in value-based argumentation, which has been proposed for practical reasoning, one needed to hand pick the arguments and the way they are related~\cite{bench-capon-value-based}, although recent work is aimed at generating them automatically \cite{abstract-argue}. 

\subsection{Proposal and novelty}

To help deal with the challenges above, we propose that we focus on the bigger picture, the larger system, the abstract process, through this first step of becoming aware of the different parts that make up the process of automated decision-making, as described in the next chapter. This allows us to then identify and address the issue from that systemic point of view, and formulate further research accordingly.

We believe that this brings multiple benefits. It should provide a more accurate reflection of human decision-making, as we discuss below, as well as more flexibility due to the modular aspect of such a system. As we shall see below, this also allows us to better deal with new scenarios and risks and also to hedge against the risks brought on by usual rule-based approaches to the problem. 

The central idea is to consider the different parts of creating a reasoner. Using this new categorisation we can get some structure to building a decision-making system that we can then base our analysis upon. Thus we can hopefully work towards a consensus on the right measurement tools and the right ways of discussing the different parts of such a system.

\section{The three R's - An Ontology} 

We thus put forward our ontology of these parts, which we call the three R’s: \textbf{Relevance}, \textbf{Representation} and \textbf{Reasoning}.
We use as a running example, an implementation in the form of a value-based moral decision-making system, MARS, as found in \cite{mars}. We now focus on the reasons behind using these steps and potential risks and failures brought about by ignoring them, in order to argue for a proactive approach to decision-making aligned with these categories.



\subsection{Relevance - identifying saliency from the context}

Imagine we are designing an autonomous car. The act of "seeing" a pedestrian in the right-hand quadrant, for instance, is a relevant aspect to the car's decision-making system, if we want to avoid collisions. Identifying this and other relevant factors is what we call the \emph{Relevance} step. This is also known as ``salience'' in image processing~\cite{grad-cam}. 

Using our running example of \cite{mars}, the Relevance step is the part in which the potential actions and the values relevant to the decision-making are given, as well as the connection between the two (which actions promote which values) with the aim of getting the agent to understand the situation it is in and the decision problem it is facing. In their case, this would be done manually by the programmer, in the set-up of a particular use of their system, but we can imagine an application like a self-driving car gathering information automatically through its vision system, for instance. 

Not performing this step at all would lead to major issues (by not caring about what, by definition, matters), and we struggle to imagine a good reasoner that does not take the context of its acting into account. If we performed this step incorrectly, perhaps its perception of the environment might not be accurate, or its comprehension of the conceptual obstacles (situations, consequences) it ought to navigate might be lacking and thus lead it to misbehave, which would give rise to various risks, some of which we have mentioned above.

Arguably, the main risk of not performing this step properly is that we may omit some important factors and thus, by not taking them into account, give rise to risks and misbehaviour. To exemplify using the self-driving car, by not taking into account the life and safety of the driver, we may end up trying to save the pedestrian but not give any thought to the driver's life (or vice versa). 

Note that we may not need to take all the relevant factors into account at the same time, or for the same decision problem, so we will need a way to narrow down the list of those we ought to act on and those that we can pass over. This will be handled in the last step where we discuss Reasoning. We also need a way to put the relevant information into a technical form, and this is our next step, Representation.


        
\subsection{Representation - syntax and semantics}
We may have various underlying implementations for our decision-making system, whether they be symbolic or logic-based systems~\cite{deontic}, based on machine learning, or otherwise. Our vision is to be implementation-agnostic, to ensure our framework can be used across various representations, applications, and platforms.  

Now, the question is how do we create a useful representation that can take in the aspects identified in the first step? This is not only a problem of a symbolic nature, namely, how to best use syntax to get the right semantics, but also one of technical relevance, because it is on the basis of whatever technical framework we will use to reason that we choose those salient aspects that we can actually use and act upon. These will depend on the use case and whatever measurement of success applies.

Some of the main issues around this step are those of conveying meaning in symbolic form. The syntax we use obviously varies with implementation, but generally we have the problem of how to best use this syntax to encompass the meaning we want the machine to grasp about the situation, the environment, the principles of acting etc. Consequently, there is the potential risk of the misinterpretation of any rules we have used, whether they be rules for action (strict or probabilistic) or rules in the sense of code used to program the machine. An extensive discussion of this issue, sometimes called the \emph{Interpretation Problem}, can be found in~\cite{interpretation}.

Note that, as we said above, some relevant aspects identified in the first step may not be used in all decisions. 
For instance, to continue with the self-driving car example, maybe we do not have effectors good enough to act upon some of the stimuli the car receives, or we do not understand the context well enough to be able to properly generate semantics (brand new elements in the configuration of the environment). The result of this second step, namely the representation built, will thus need to connect the relevant aspects of the situation with the reasoning part while attempting to leverage the underlying implementation as well as possible. Therefore, the core idea of this step is to create a representation that is abstract enough to cover many situations but flexible enough to be amenable to being used in new cases and powerful enough (expressive enough, for instance, if it is a type of logic) to proactively address potential risks not previously encountered.

For instance, in our running example of MARS, moral paradigms are used as the representation, and they compile an ordered list of values that are important to the agent, in the form of strata. This representation works because it plays well with the underlying technical implementation (the programming) as well as the abstract language used (the logic used for formal reasoning) \cite{mars}. For a self-driving car, these values may be human life, safety, no component failure, the rules of the road, minimizing time to destination, etc. The importance of these considerations varies based on the context in which they are applied, so this is why the former step, Relevance, is so important, as it sets up the particulars of the decision problem at hand.


As we have mentioned before, a key to third wave autonomous systems is context adaptation. Thus, these values could have a changing relative importance depending on the active components of the vehicle. For instance, at high speeds, safety is of utmost importance. On the freeway, the rules of the road are minimal. In suburban environments, we may want to ensure a very low tolerance for misidentification of potential pedestrians. The representation should be flexible enough to allow for this.
            
\subsection{Reasoning} 
 
Now, having identified the relevant aspects to the decision-making process, and having built a useful representation that can take them into account, we come to the step of actually making decisions based on the acquired information.

Arguably, the issues of highest importance here and which may give rise to unexpected risks are quick decision making and confidence, with the aims of minimising false positives and false negatives as well as building a sense of trust in our systems. This is essential for numerous practical reasons: for our system to be adopted, pass regulatory requirements, be acquired by the end user, be used appropriately etc.

Furthermore, and perhaps the most difficult issue of doing this step, is choosing what rules and principles to have the machine follow. The field of \emph{value alignment} considers this issue of getting the values we desire into the behaviour and reasoning of our creations, but even there, most attempts to achieve value alignment aim to implement the same rules we seem to follow into the machine~\cite{taylor}~\cite{soares}. Thus, a further potential issue is whether we need a descriptive or normative approach. That is, should we implement the rules we currently use ourselves with our potential biases (descriptive) or ones we would ideally want to follow as rational decision-makers (normative)?
 
 
We strongly believe that the most important point and why this step is essential is that reasoning is the key component for anticipatory thinking. A problem frequently discussed in the literature is that some reasoning systems have rigid rule-based frameworks, and this leads to an inability to obtain the kind of flexibility, coverage and abstraction required to properly reason under uncertainty in open environments, which could, we hope, be improved by taking into account the three steps discussed here. Consider, for instance, \emph{rule explosion}, having to add more and more rules to deal with new cases that the system was not programmed to deal with. Or consider how, in terms of moral reasoning, virtue ethics could be the best moral paradigm in certain use cases \cite{cosmin-virtue-ethics}, whereas consequentialism or deontology may be desired in others \cite{cons}. Another issue is the risk of being stuck in local minima, and thinking the solution is optimal when it is not, which happened, for instance, in DeepMind's Atari games playing agent~\cite{atari} which struggled due to greedy algorithms. 

There are potential risks around not handling this step right. We may get decision paralysis, as, for instance, it is described in the paradox of Buridan and his donkey~\cite{lamport-buridan}) where one cannot rationally decide between options; or we may have the opposite, an over-active decision-maker, potentially arising, for instance, in a self-driving car when a collision with a small object is imminent and, as driver, it may be advisable to hold the course instead of risking a more serious accident by attempting to perform any other mitigation.


As an example of how to approach this step, in MARS they have multiple different flavours of reasoners (models) that combine the previously identified relevant points and represented values into making a decision that is arguably in tune with the desired outcome, given a particular value ordering (moral paradigm). For instance, they have an additive model that adds up the values involved in actions to select the right action to perform, a weighted model that assigns numerical weights to the values based on their relative importance, as well as other more refined ones. They argue that these different models allow for flexible decision-making. 
     
\section{Contribution and Discussion}


Two essential points of this proposal are \emph{adaptability} and \emph{redundancy}: there are multiple ways to reason about a problem, and we might not be certain of the best one \emph{a priori}. If one reasoning framework is inadequate or wrong, then there may be another way to solve the problem, and we should be able to account for this, and our modular approach should allow us to shift the relative importance of the relevant features, with flexibility in what paradigm we use. Our approach aims to support third wave autonomous systems, in which the values themselves may be re-prioritized in different circumstances, as well as the reasoning paradigm changed completely. 

In the Relevance step, we ought to be able to adapt which features are salient based on the context, in the Representation step, the structures we build and feed into the reasoning need to be adaptable, and in the Reasoning step the paradigm or rules we use should be applicable in most cases and flexible enough to give desired outputs. The separation into the three R's aims to make it easier to reason about different components by enforcing cohesion and reducing coupling as well as to reduce computational complexity in the implementation. 

 
In conclusion, in this paper, we contribute to the solving of potential risks that arise in anticipatory systems, by proposing a nomenclature that reflects the literature on decision-making, and an ontology that allows researchers that adopt it to clarify the focus of their work. We hope to improve future work around this and adjacent areas through this shared framework.
 


\bibliography{main}

\begin{thebibliography}{}

\bibitem[\protect\citeauthoryear{Badea and Artus}{2021}]{interpretation}
Badea, C., and Artus, G.
\newblock 2021.
\newblock Morality, machines and the interpretation problem: A value-based,
  wittgensteinian approach to building moral agents.
\newblock {\em arXiv preprint arXiv:2103.02728}.

\bibitem[\protect\citeauthoryear{Badea}{2020}]{mars}
Badea, C.
\newblock 2020.
\newblock Have a break from making decisions, have a mars: The multi-valued
  action reasoning system.
\newblock {\em arXiv preprint arXiv:2109.03283}.

\bibitem[\protect\citeauthoryear{Bench-Capon}{2003}]{bench-capon-value-based}
Bench-Capon, T.~J.
\newblock 2003.
\newblock Persuasion in practical argument using value-based argumentation
  frameworks.
\newblock {\em Journal of Logic and Computation} 13(3):429--448.

\bibitem[\protect\citeauthoryear{Boureau, Sokol-Hessner, and
  Daw}{2015}]{medical}
Boureau, Y.-L.; Sokol-Hessner, P.; and Daw, N.~D.
\newblock 2015.
\newblock Deciding how to decide: Self-control and meta-decision making.
\newblock {\em Trends in cognitive sciences} 19(11):700--710.

\bibitem[\protect\citeauthoryear{Cabrio and Villata}{2012}]{abstract-argue}
Cabrio, E., and Villata, S.
\newblock 2012.
\newblock Generating abstract arguments: A natural language approach.
\newblock In {\em COMMA},  454--461.

\bibitem[\protect\citeauthoryear{Dietrich and
  List}{2013}]{preferences-come-from}
Dietrich, F., and List, C.
\newblock 2013.
\newblock Where do preferences come from?
\newblock {\em International Journal of Game Theory} 42(3):613--637.

\bibitem[\protect\citeauthoryear{Hindocha and
  Badea}{2021}]{cosmin-virtue-ethics}
Hindocha, S., and Badea, C.
\newblock 2021.
\newblock Moral exemplars for the virtuous machine: the clinician’s role in
  ethical artificial intelligence for healthcare.
\newblock {\em AI and Ethics}  1--9.

\bibitem[\protect\citeauthoryear{Jones and Sergot}{1996}]{deontic}
Jones, A.~J., and Sergot, M.
\newblock 1996.
\newblock A formal characterisation of institutionalised power.
\newblock {\em Logic Journal of the IGPL} 4(3):427--443.

\bibitem[\protect\citeauthoryear{Lamport}{2012}]{lamport-buridan}
Lamport, L.
\newblock 2012.
\newblock Buridan’s principle.
\newblock {\em Foundations of Physics} 42(8):1056--1066.

\bibitem[\protect\citeauthoryear{Launchbury}{2017}]{DARPA}
Launchbury, J.
\newblock 2017.
\newblock A darpa perspective on artificial intelligence.
\newblock {\em Retrieved November} 11:2019.

\bibitem[\protect\citeauthoryear{Mnih \bgroup et al\mbox.\egroup
  }{2013}]{atari}
Mnih, V.; Kavukcuoglu, K.; Silver, D.; Graves, A.; Antonoglou, I.; Wierstra,
  D.; and Riedmiller, M.
\newblock 2013.
\newblock Playing atari with deep reinforcement learning.
\newblock {\em arXiv preprint arXiv:1312.5602}.

\bibitem[\protect\citeauthoryear{Pang \bgroup et al\mbox.\egroup
  }{2020}]{local-anomaly}
Pang, G.; Shen, C.; Cao, L.; and van~den Hengel, A.
\newblock 2020.
\newblock Deep learning for anomaly detection: {A} review.
\newblock {\em CoRR} abs/2007.02500.

\bibitem[\protect\citeauthoryear{Selvaraju \bgroup et al\mbox.\egroup
  }{2017}]{grad-cam}
Selvaraju, R.~R.; Cogswell, M.; Das, A.; Vedantam, R.; Parikh, D.; and Batra,
  D.
\newblock 2017.
\newblock Grad-cam: Visual explanations from deep networks via gradient-based
  localization.
\newblock In {\em 2017 IEEE International Conference on Computer Vision
  (ICCV)},  618--626.

\bibitem[\protect\citeauthoryear{Soares}{2016}]{soares}
Soares, N.
\newblock 2016.
\newblock The value learning problem.
\newblock In {\em Ethics for Artificial Intelligence Workshop at IJCAI-16}.

\bibitem[\protect\citeauthoryear{Taylor \bgroup et al\mbox.\egroup
  }{2016}]{taylor}
Taylor, J.; Yudkowsky, E.; LaVictoire, P.; and Critch, A.
\newblock 2016.
\newblock Alignment for advanced machine learning systems.
\newblock {\em Ethics of Artificial Intelligence}  342--382.

\bibitem[\protect\citeauthoryear{Wang}{2000}]{meta}
Wang, Z.
\newblock 2000.
\newblock Meta-decision making: concepts and paradigm.
\newblock {\em Systemic practice and action research} 13(1):111--115.

\bibitem[\protect\citeauthoryear{Williams and Yampolskiy}{2021}]{ai-failures}
Williams, R., and Yampolskiy, R.
\newblock 2021.
\newblock Understanding and avoiding ai failures: A practical guide.
\newblock {\em Philosophies} 6(3).

\bibitem[\protect\citeauthoryear{Yampolskiy}{2018}]{cons}
Yampolskiy, R.~V.
\newblock 2018.
\newblock {\em Artificial intelligence safety and security}.
\newblock CRC Press.

\end{thebibliography}
\bibliographystyle{aaai}

\end{document}